\documentclass[lettersize,journal]{IEEEtran}
\usepackage{amsmath,amsfonts}
\usepackage{algorithmic}
\usepackage{algorithm}
\usepackage{array}
\usepackage[caption=false,font=normalsize,labelfont=sf,textfont=sf]{subfig}
\usepackage{textcomp}
\usepackage{stfloats}
\usepackage{url}
\usepackage{verbatim}
\usepackage{graphicx}
\usepackage{cite}
\usepackage{multirow}
\usepackage{booktabs}
\usepackage{caption}
\usepackage{color}
\definecolor{red}{rgb}{0,0,0}

\begin{document}

\title{\huge{LACV-Net: Semantic Segmentation of Large-Scale Point Cloud Scene via Local Adaptive and Comprehensive VLAD}}

\author{{Ziyin Zeng, Yongyang Xu, Zhong Xie, Wei Tang, Jie Wan and Weichao Wu}

\thanks{Ziyin Zeng, Yongyang Xu, Zhong Xie and Wei Tang are with School of Computer Science, China University of Geosciences, Wuhan, China. (email: yongyangxu@cug.edu.cn)

Jie Wan and Weichao Wu are with Key laboratory of geological and evaluation of ministry of education, China University of Geosciences, Wuhan, China.}}

\maketitle

\begin{abstract}
Large-scale point cloud semantic segmentation is an important task in 3D computer vision, which is widely applied in autonomous driving, robotics, and virtual reality. Current large-scale point cloud semantic segmentation methods usually use down-sampling operations to improve computation efficiency and acquire point clouds with multi-resolution. However, this may cause the problem of missing local information. Meanwhile, it is difficult for networks to capture global information in large-scale distributed contexts. To capture local and global information effectively, we propose an end-to-end deep neural network called LACV-Net for large-scale point cloud semantic segmentation. The proposed network contains three main components: 1) a local adaptive feature augmentation module (LAFA) to adaptively learn the similarity of centroids and neighboring points to augment the local context; 2) a comprehensive VLAD module (C-VLAD) that fuses local features with multi-layer, multi-scale, and multi-resolution to represent a comprehensive global description vector; and 3) an aggregation loss function to effectively optimize the segmentation boundaries by constraining the adaptive weight from the LAFA module. Compared to state-of-the-art networks on several large-scale benchmark datasets, including S3DIS, Toronto3D, and SensatUrban, we demonstrated the effectiveness of the proposed network.
\end{abstract}

\begin{IEEEkeywords}
Point Cloud, Semantic Segmentation, Large-Scale Scene, Local and Global Feature
\end{IEEEkeywords}

\section{Introduction}
With the development of 3D sensor technology, 3D point cloud data are widely used in realistic scenarios. Recently, researches have successfully applied small-scale point clouds to tasks such as face recognition \cite{3Dface_1,3Dface_2,3Dface_3}, hand pose reconstruction \cite{3Dhand_1,3Dhand_2,3Dhand_3} and human pose estimation \cite{3Dhumanpose_1,3Dhumanpose_2,3Dhumanpose_3}. Furthermore, large-scale point cloud scene semantic segmentation has been widely focused on as an important task in 3D scene understanding, and has significant potential in applications such as autonomous driving, robotics and virtual reality \cite{PPN,yang,DGANet}. 

Although some successful researches, such as Recurrent Neural Networks (RNNs) and Convolutional Neural Networks (CNNs), have performed significantly in fields of ordered and structured 1D natural language processing and 2D image processing, they are difficult to apply in disordered and unstructured 3D point clouds processing \cite{PointNet}. Particularly, large-scale point cloud scene is collected with a considerable number of points (millions or even billions); therefore, semantic segmentation of large-scale point cloud scenes is a challenge.

To efficiently process large-scale point cloud, recent studies use an encoder–decoder network framework with down-sampling operations \cite{PointNet++,RandLA-Net,ResDLPS-Net,AD-SAGC,BAAF-Net,PointASNL,SCF-Net,BAF-LAC}. Therefore, the point cloud is down-sampled after each encoding layer, and the number of points processed by encoding layers can be reduced, which saves computational resources and suitable for large-scale point cloud processing. \textcolor{red}{However, the down-sampling operation may lose local details \cite{RandLA-Net}. In addition, the contextual information of large-scale point cloud scenes is always distributed over a large spatial scale \cite{AD-SAGC}. The network has difficulty in capturing the distributed contexts in a large-scale scene. Therefore, how to effectively learning local and global information is a crucial issue in large-scale point cloud semantic segmentation.} In this study, we use an encoder-decoder network framework to directly learn fine-grained point features, while we endeavor to address some of the drawbacks of existing studies on local and global feature representations:

\textbf{Local perception ambiguity} Some recent studies represent the local features of point clouds based on their pre-defined neighbors, such as combining features (geometric or semantic) from neighboring points with the centroid \cite{RandLA-Net,BAAF-Net,SCF-Net,BAF-LAC,DLA-Net}. This is efficient computation and suitable for large-scale point cloud scenes without sophisticated algorithms. However, centroids perceive the neighbor consistently during the neighbor’s construction, which may cause segmentation errors for points distributed near the boundaries of different semantic classes (as shown in Fig.3). To alleviate possible impacts, and in particular, to make the centroid focus more on significant neighboring points instead of each neighboring point equally, we design a robust local information encoding unit to learn the local context and make full use of the local neighbor information, and introduce an adaptive augmentation unit to learn the adaptive weight, which represents the centroid perception on the local neighbor based on the difference between features of the centroid and its neighbor. Moreover, an aggregation loss function is proposed to effectively optimize the segmentation boundaries by constraining the adaptive weight such that neighboring points with smaller differences would have more influence on the centroid, which also accelerates the convergence of the network. 

\textbf{Global feature insignificant} \textcolor{red}{Most recent studies on learning global features use global max or average pooling to capture global descriptors \cite{PointNet,DGCNN,GMaxPool_1,PCT}. However, most features are lost in the pooling process \cite{NetVLAD}, resulting in insufficient acquired global feature mapping for fine-grained semantic segmentation.} Some studies use global attention to obtain global information \cite{PointASNL,AD-SAGC,PCT}; however, when the number of points in per-batch is enormous, attention-based methods are computationally expensive and memory inefficient. VLAD module \cite{VLAD,NetVLAD} has been demonstrated superior performance in 2D position recognition by capturing global features, and some studies \cite{PointNetVLAD,VLADpoint_1,VLADpoint2} in 3D point cloud place recognition use the VLAD module to capture global features with excellent results. Inspired by them, we designed a comprehensive VLAD module (C-VLAD) to efficiently acquire global information from 3D point clouds by fusing local features with multi-layer, multi-scale, and multi-resolution. Compared with global max or average pooling, the proposed C-VLAD module can capture more distinguishable global information for fine-grained semantic segmentation.\\

In summary, our key contributions are: \\

\begin{itemize}

\item We propose a \textbf{local adaptive feature augmentation module (LAFA)} to learn local adaptive weight and adequately represent the local features.\\

\item We introduce a \textbf{comprehensive VLAD module (C-VLAD)} to capture distinguishable fine-grained global descriptors.\\

\item We design an \textbf{aggregation loss function} to constrain the adaptive weight from the LAFA module, and accelerates the convergence and fitting of the network.\\

\item We evaluate our network on three large-scale benchmarks. The experimental results demonstrate that our method outperforms state-of-the-art methods.

\end{itemize}

\section{Related Work}

\textcolor{red}{In this section, the two main related techniques: learning-based method on point cloud, and local and global feature representation in segmentation are discussed.}

\subsection{Learning-based method}

The goal of semantic segmentation is to assign per-point semantic label, which accurately describes the type of object in scene, and helps to achieve scenario understanding. Traditional methods typically employ hand-crafted features \cite{tradition_1,tradition_2}. In recent years, deep learning has been applied to point cloud understanding and achieve excellent performance. Recent learning-based methods can be mainly divided into projection-based, voxel-based, and point-based methods, which are discussed as follows.

Since point clouds are disordered and unstructured, it is difficult to apply widely used CNNs to 3D point cloud data. Projection-based \cite{PPN,PB_1,PB_2,PB_3} and voxel-based \cite{VB_1,VB_2,VB_3} methods convert unstructured point cloud data into structured 2D images or 3D voxels by pre-processing, while apply successful CNNs to process these structured data. However, these methods may cause the problem of losing detailed information in projecting or voxelizing, while the computation cost is expensive, especially when processing large-scale point cloud data. Point-based methods are designed to directly process point cloud without using intermediate variants (projections or voxels). The pioneering PointNet \cite{PointNet} directly uses point-wise multi-layer perceptions ($mlps$) to learn per-point features. Following this successful work, an increasing number of point-based methods have been proposed to capture patterns lying in the points. \cite{PointNet++,AD-SAGC,KPConv,RandLA-Net,BAAF-Net,PointWeb,PointASNL,SCF-Net,BAF-LAC,leard}. Some recent works have applied RNNs \cite{3P-RNN,RSNet}, Graph Neural Networks (GNNs) \cite{SPG,GACNet,DeepGCNs,RG-GCN}, and Transformer \cite{PCT,PT,PointASNL}, which are mature applications in other applications, to point cloud semantic segmentation with superior results. Although some complex methods achieve excellent performance in point cloud understanding, they are not applicable to large-scale point cloud scene semantic segmentation due to their excessive computational cost. In this study, we only use $mlps$ with remarkable memory and computational efficiency to connect neural units.

\subsection{Local and global feature representation in segmentation}

For local feature representation, PointNet \cite{PointNet} uses point-wise $mlps$ to learn per-point features; thus, it suffers from many shortcomings in local details. Different from PointNet, the following works design various local aggregation modules to efficiently focus on feature representation in local neighbor \cite{PointNet++,DGCNN,ShellNet,PointGCR,PointConv_CE,FPC,leard}. Some recent studies combine the features of the neighbor with the centroid, they have efficient computation and suitable for large-scale point cloud scenes \cite{RandLA-Net,BAAF-Net,SCF-Net,BAF-LAC,DLA-Net}. For example, RandLA-Net \cite{RandLA-Net} combines the spatial coordinates of neighboring points and centroid to explicitly represent the local geometric structure, BAAF-Net \cite{BAAF-Net} combines both semantic features and spatial coordinates of neighboring points and centroids to augment the local context. However, these methods percept of each centroid on the neighbors consistently in the process of neighboring construction, which may cause segmentation errors in points near the boundaries of different semantic classes. We design a local adaptive feature augmentation module to effectively avoid the problem of local perception ambiguity, by learning the influence of neighboring points on the centroid.

Current studies typically use global max or average pooling to capture global features from the previously acquired local features \cite{PointNet,DGCNN,GMaxPool_1,PCT}. However, using global max or average pooling loses substantial features, resulting in insufficient global feature mapping for fine-grained semantic segmentation. Some methods use global attention to obtain global information \cite{AD-SAGC,PointASNL,PCT}, but when the number of points in per-batch is enormous, the attention-based methods are computationally expensive and memory inefficient. The VLAD module \cite{VLAD,NetVLAD} performs well in 2D image place recognition owing to its effective acquisition of global descriptors from the clustering of captured local features, and some methods \cite{PointNetVLAD,VLADpoint_1,VLADpoint2} have used the VLAD module for point cloud place recognition with excellent results. In this paper, a C-VLAD module is proposed to fuse multi-layer, multi-scale and multi-resolution local features and learn distinguishable global information to capture comprehensive representations. 

\section{Method}

\textcolor{red}{In this section, the details of the proposed network: LACV-Net are presented. The LACV-Net consists of three core modules: 1) Local adaptive feature augmentation module (LAFA) to adequately represent the local features mappings, 2) Comprehensive VLAD module (C-VLAD) for capturing long range contextual information globally, 3) Aggregation loss function for accelerating the convergence and fitting of the network. The descriptions of the proposed network and each module are detailed as follows.}

\subsection{Network architecture}

\begin{figure*}[!t]
\vspace{-1.5em}
\includegraphics[width=1.0\textwidth]{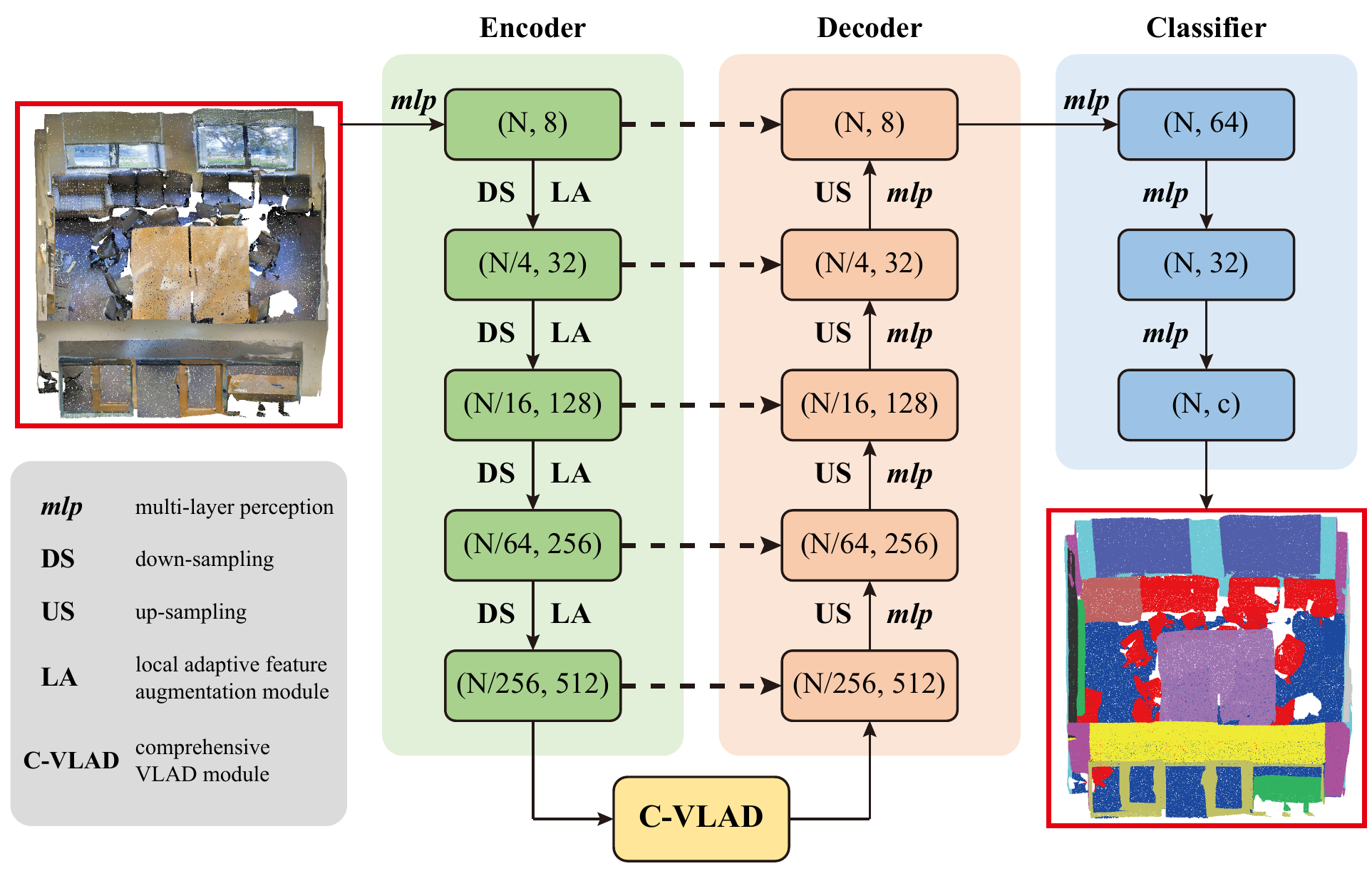}
\caption{Architecture of the proposed network.}
\label{}
\vspace{-1.5em}
\end{figure*}

Fig.1 shows the detailed architecture of the proposed network, which follows the encoder-decoder framework. Each encoding layer contains a local adaptive feature augmentation module (LAFA) and a down-sampling operation. We use random sampling operations to down-sample the point cloud because of its computational efficiency, which is suitable for large-scale point cloud scene data. In this aspect, this behavior of processing 3D point clouds is similar to that of traditional CNNs for 2D images, which increases the number of channels while simultaneously narrowing the image size to provide a concise description. Between the encoder and decoder, we added a C-VLAD module that fuses the local features acquired in all the previous encoding layers. Each decoding layer contains an up-sampling operation and $mlp$. Features are transferred between the encoding and decoding layers based on skip connections. Finally, in the classifier, three multi-layer perceptions are used as fully-connected layers, as well as a softmax layer to predict label classification scores for each point. The semantic segmentation result is determined by the label with the highest score.

\subsection{Local adaptive feature augmentation module}

Recent studies typically employ networks with down-sampling operations as mentioned before. Since down-sampling results in loss of local information, local information extraction is critical for accurate semantic segmentation of point clouds. Some information extraction modules \cite{RandLA-Net,BAAF-Net,SCF-Net,BAF-LAC,DLA-Net} mainly combine the information of the $K$ neighbor points with the centroid to improve the perception of the model. They work excellently in the task of large-scale point cloud semantic segmentation because of efficient computing rather than complex convolution.

However, these methods have the following drawbacks: 1. They mainly focus on encoding the spatial coordinates and semantic features of point clouds in each encoding layer. But the color information $rgb$ is only input to the network as an initial feature and not encoded in each encoding layer like spatial coordinates or semantic features, which can significantly improve the accuracy of semantic segmentation. 2. Each neighboring point's influence on the centroid is equal in the process of neighbor construction; however, the perception of centroids on neighbor features is ambiguity, and the centroids should focus on the more significant points rather than equally on each point in the local neighbor. To address these problems, as shown in Fig.2, we design a local adaptive feature augmentation module (LAFA) module. The proposed LAFA module contains two neural units: 1) a local information encoding unit, and 2) an adaptive augmentation unit.

\begin{figure*}[!t]
\centering
\includegraphics[width=1.0\textwidth]{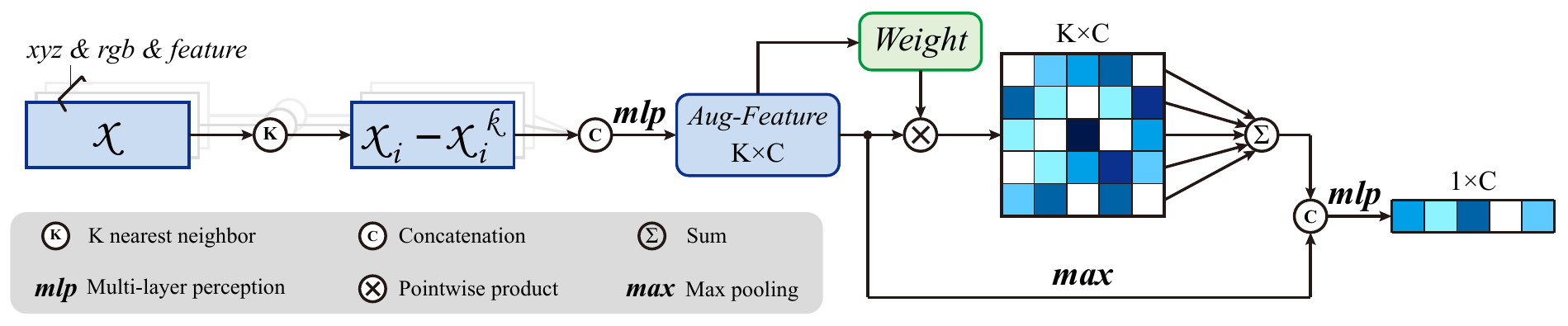}
\caption{Proposed local adaptive feature augmentation module (LAFA).}
\label{}
\vspace{-1.0em}
\end{figure*}

\subsubsection{Local information encoding unit}

Available datasets of point clouds usually contain spatial coordinates $xyz$ and color information $rgb$, which are intrinsic information of the point clouds. Existing studies of local feature encoding focus solely on point spatial coordinates $xyz$ and semantic features $f$, resulting in underutilization of color information $rgb$. The color information $rgb$ is only input to the network as an initial feature and not encoded in each encoding layer like spatial coordinates $xyz$. In our implementation, for each point $p_i$ in point cloud $P$, the spatial geometry, color information and semantic features of the nearest $K$ points, which are determined by the K-nearest neighbor algorithm (KNN) based on the Euclidean distance in the 3D space, are fused into the central point, ensuring that the context information is gathered.

We only use the relative difference between the centroid and its neighbor to encode the local spatial information as $\ell(p_i)=mlp(p_i - p_i^k)$. Similarly, local color information encoding can be represented as $\ell(c_i)=mlp(c_i -  c_i^k)$, and local semantic information encoding can be represented as $\ell(f_i)=mlp(f_i - f_i^k)$. Finally, we concatenate the local spatial information, local color information and local semantic information to obtain the local information encoding $\triangle \ell\in\mathbb{R}^{N\times K\times C}$ as follows:

{\setlength\abovedisplayskip{1.5pt}
\begin{align}
\triangle \ell_i=\{\ell(p_i)\oplus\ell(c_i)\oplus\ell(f_i)\}
\end{align}
}

\noindent The local information encoding $\triangle \ell_i$ is only combined by the relative difference of spatial geometry, color information and semantic features, so it can effectively represent the feature difference between neighboring points and the centroid.

\subsubsection{Adaptive augmentation unit}

\begin{figure}[!t]
\centering
\includegraphics[width=0.45\textwidth]{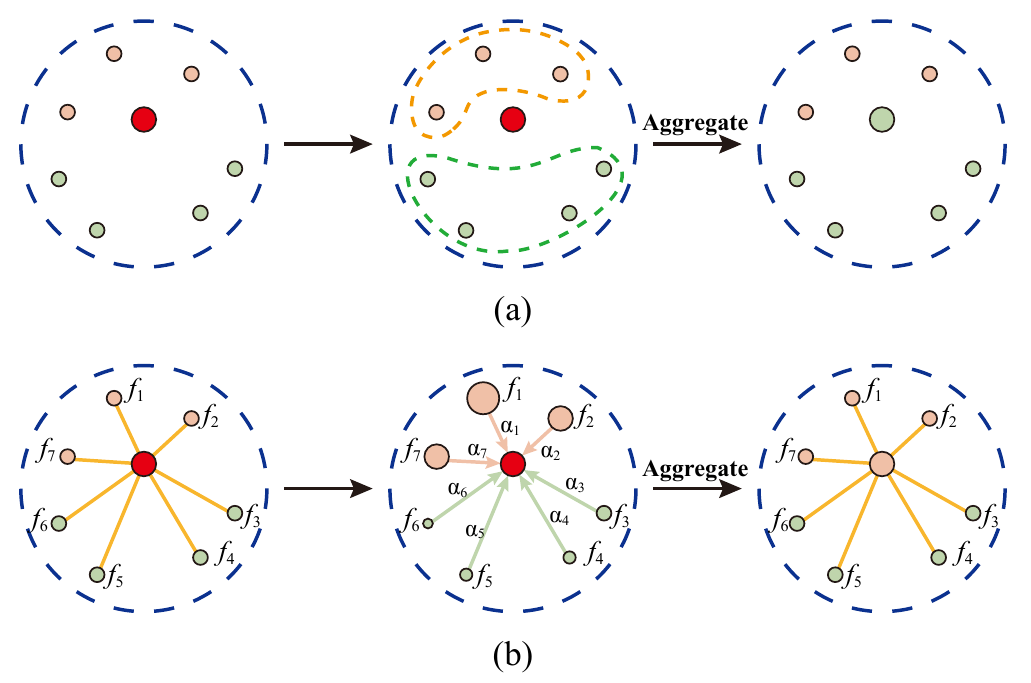}
\caption{Schematic with and without adaptive weight. The Fig.3-(a) indicates without adaptive weighting unit, and the Fig.3-(b) indicates with adaptive weighting unit.}
\label{}
\vspace{-1.0em}
\end{figure}

As shown in Fig.3-(a), when points are distributed near semantic boundaries, simply aggregating features of neighboring points of the centroid may cause the problem of incorrect boundary segmentation. To solve the problem of ambiguous local perception and fully utilize the local information while benefiting the network to segment the boundaries accurately, as shown in Fig.3(b), we learn the adaptive weight of point clouds by calculating the similarity between the local information of neighboring points and the centroid.

We define similarity matrix $S$ as the feature difference between neighboring points and the centroid. Because the local information encoding $\ell_i$ can effectively represent the feature difference, we define the similarity matrix $S\in\mathbb{R}^{N\times K\times C}$ by linear mapping of feature differences in local information as follows: 

{\setlength\abovedisplayskip{1.5pt}
\begin{align}
S=\mathcal{M}({\triangle \ell_i})
\end{align}
}

Where $\mathcal{M}$ is linear mapping. Because points with more similar features are more likely to be classified into one class, we normalized the similarity matrix $S$ using the $softmax$ function in the neighbor dimension to obtain the adaptive weight $W$. The adaptive weight $W\in\mathbb{R}^{N\times K\times C}$ are expressed as follows:

{\setlength\abovedisplayskip{1.5pt}
\begin{align}
W = \frac{exp(S^k)}{\sum_{k=1}^{K}exp(S^k)}
\end{align}
}

Finally, we multiply the adaptive weight $W$ by the local information encoding $\triangle\ell$ to obtain the weighted local information encoding. We use the $sum-pooling$ operation in the neighbor dimension to aggregate the entire local neighbor, and use the $max-pooling$ operation to capture the local significant features. The adaptive local features $F\in\mathbb{R}^{N\times 1\times C}$ can be represented as follows:

{\setlength\abovedisplayskip{1.5pt}
\begin{align}
F=mlp\{(\sum_{k=1}^{K}{(W\triangle\ell)}_k) \oplus (max{(\triangle\ell)}_k)\}
\end{align}
}

\subsection{Comprehensive VLAD module}

\begin{figure*}[!t]
\centering
\includegraphics[width=1.0\textwidth]{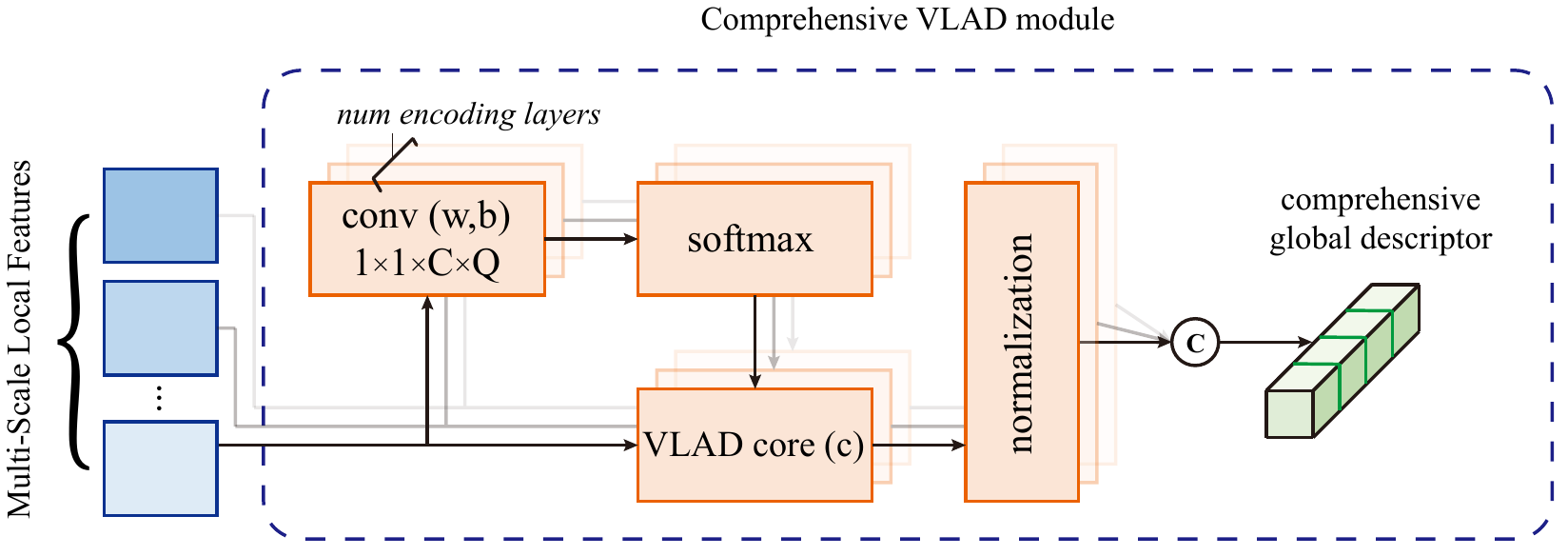}
\caption{Proposed comprehensive VLAD module (C-VLAD).}
\label{}
\vspace{-1.0em}
\end{figure*}

The original VLAD module \cite{VLAD,NetVLAD} is designed to aggregate local features of images learned from hand-crafted or VGG \cite{VGG}/AlexNet \cite{AlexNet} into a global descriptor vector. Typically, the local features input to the VLAD module are output of the latest layer, as implemented in NetVLAD \cite{NetVLAD}. To efficiently analyze a real 3D scene consisting of a large number of points, we use down-sampling operations to reduce the number of points in each encoding layer, which corresponds to decrease resolution in 2D image processing. Although it can be easily realized by applying the proposed LAFA modules for down-sampled point cloud subsets, the corresponding output features become implicit and abstract. Therefore, we designed a comprehensive VLAD module for fusing local features with multi-layer, multi-scale and multi-resolution as follows.

The multi-scale local features $\{f_1, f_2 ..., f_L | f_l\in\mathbb{R}^{N_l\times C_l}\}$ are obtained after multiple encoding layers, where $f_l$ is the local feature output from the $l$-$th$ encoding layer, $N_l$ is the number of points in the $l$-$th$ layer before the point cloud is downsampled, and $C_l$ is the feature dimension of the $l$-$th$ layer. The VLAD module learn $Q_l$ cluster centers $\{c_{l,1}, c_{l,2} ..., c_{l,Q} | c_{l,q}\in\mathbb{R}^{N_l\times C_l}\}$. Subsequently, the subvector $v_{l, q}$ (the $q$-$th$ cluster center for the $l$-$th$ layer's local features) is calculated for each cluster center $c_{l,q}$ as follows:

{\setlength\abovedisplayskip{1.5pt}
\begin{align}
\vspace{-1.0em}
v_{l,q}=\frac{exp({w_{l,q}^Tf_l+b_{l,q}})}{\sum_{q}exp({w_{l,q}^Tf_l+b_{l,q}})}(f_l-c_{l,q})
\end{align}
}

Each global descriptor vector $v$ corresponding to each local feature of the layer can be expressed as $v_l = \{v_{l, 1}, v_{l, 2}, …, v_{l, Q_l}\}$, where $v_l \in\mathbb{R}^{(Q_l\times C_l)}$. Then we concatenate all output global descriptor vectors to fuse local features with multi-layer, multi-scale and multi-resolution, and obtain the comprehensive global descriptor vector $V \in\mathbb{R}^{(L\times Q_l\times C_l)}$ as follows: 

{\setlength\abovedisplayskip{1.5pt}
\begin{align}
\vspace{-1.0em}
V = concat(v_1, v_2, ..., v_l)
\end{align}
}

Compared to the traditional VLAD module that only takes the output of the latest layer as input features, which cannot fully utilize the local information of different layers, our C-VLAD computes the local information of each layer in parallel and aggregates the local information with multi-layer, multi-scale and multi-resolution to capture a comprehensive global information.

\subsection{Aggregation loss function}

In this section, we first describe the design of constraint loss function $\mathcal{L}(f_i)$.  We want to further optimize the adaptive weight that better represent the influence of the neighboring points on the centroids, such that the points in the vicinity of the boundary can be correctly classified. 

Since the adaptive weight are computed from the local information differences between the neighboring points and the centroid, and points with more similar features are more likely to be classified into the same class, we expect the smaller the difference between neighboring and centroid features is, the more significant it is on the adaptive weight. We consider the adaptive weighted local information as a feature offset, and we encourage the shifted neighboring features to be closer to the centroid by minimizing the $L_1$ distance:

{\setlength\abovedisplayskip{1.5pt}
\begin{align}
\mathcal{L}(f_i)=\left|\left| f_i - \sum_{k=1}^{K}(f_i^k + {W_i^k\triangle\ell_i^k}) \right|\right|_{L_1}
\end{align}
}

This is called constraint loss function. Here we do not use the generalized $L_2$ regularization, mainly because the channel dimension of the features is large and the $L_2$ regularization requires more computational resources, so we use the $L_1$ regularization which has less computational overhead. We experimentally demonstrate that the proposed constrain loss function can accelerate the convergence of the model. \textcolor{red}{Generally, the weighted cross-entropy loss $\mathcal{L}_{wce}=-\sum_{c=1}^{C}w_c*(y_c\ log\ \hat{y}_c)$ is computed as back propagation and balancing different classes.} In addition, we also include the designed constraint loss $\mathcal{L}(f_i)$. Hence, for the proposed network containing $N$ encoding layers, the aggregation loss is:

{\setlength\abovedisplayskip{1.5pt}
\begin{align}
\mathcal{L}_{agg}=\mathcal{L}_{wce}+\frac{1}{N_e}\sum_{n=1}^{N}\mathcal{L}(f_i)
\end{align}
}

\noindent \textcolor{red}{Where $w_c={N}/{(n_c+0.02N)}$, $N_e$ is the number of encoding layers, $C$ is the number of class, $y_c$ and $\hat{y}_c$ denote the true label vector and predicted label vector of the $c_{th}$ class respectively, $N$ represents the number of total points, and $n_c$ is the number of points on the $c_{th}$ class. }

\section{Experiments}

\textcolor{red}{In this section, experiments have been conducted to verify the effectiveness of the proposed LACV-Net in 3D point cloud semantic segmentation task. The evaluations are performed on both indoor and outdoor large-scale benchmarks. After that, the ablation analysis on each key components of LACV-Net are provided.}

\subsection{Experimental setting and details}

To fully evaluate the effectiveness of the proposed network in point cloud semantic segmentation, we used several large-scale benchmark datasets, including the indoor dataset S3DIS \cite{s3dis} and the outdoor dataset Toronto3D \cite{Toronto3D} and SensatUrban \cite{SensatUrban}. To quantitatively analyze the performance of the proposed network, per-class intersection over union (IoUs), mean IoU (mIoU), and overall accuracy (OA) are used as evaluation metrics as follows. 

{\setlength\abovedisplayskip{1.5pt}
\begin{align}
IoU=\frac{TP}{TP+FP+FN}
\end{align}
}

{\setlength\abovedisplayskip{1.5pt}
\begin{align}
mIoU=\frac{\sum_{i=1}^{n}{IoU}_i}{n}
\end{align}
}

{\setlength\abovedisplayskip{1.5pt}
\begin{align}
OA=\frac{TP}{N}
\end{align}
}

Where $TP$ represents the number of true positive samples, $FP$ represents the number of false positive samples, $FN$ represents the number of false negative samples, $n$ represents the number of semantic classes and $N$ represents the number of total samples. The effectiveness and rationality of each module are demonstrated by ablation studies. To train the proposed network, the number of points initially input to the network is $10\times2^{12}$, batchsize is set to 6 with 100 epochs, Adam optimizer is used with default parameters to optimize the loss function, initial learning rate is set to 0.01, and neighbor search range $K$ is set to 16. All the experiments are performed on a single NVIDIA GeForce RTX3090 GPU and an i7-12700K CPU. 

During training step, in the down-sampling process, we use efficient random down-sampling to improve the computational efficiency of the network instead of using the farthest point sampling with higher accuracy but expensive computational. Particularly, the network gradually processes the lower resolutions of the point cloud as: ($N\rightarrow\ \frac{N}{4}\rightarrow\ \frac{N}{16}\rightarrow\ \frac{N}{64}\rightarrow\ \frac{N}{256}$), whereas the channel dimension of the features increases as: ($8\rightarrow\ 32\rightarrow\ 128\rightarrow\ 256\rightarrow\ 512 $). There are two consecutive LAFA modules in each encoding layer, where the output dimension of the first LAFA module is half of that of the encoding layer. Since the encoder of the network has five encoding layers and outputs five local features with multi-layer, multi-scale and multi-resolution, we use the C-VLAD module to fuse these local features to capture global descriptors, and the number of cluster centers of the C-VLAD module is set to 16. For the sake of fairness, the chosen datasets all have color information, and the chosen methods use the color information as the initial input feature.

\subsection{Dataset description}

\subsubsection{S3DIS}

The Stanford Large-Scale 3D Indoor Spaces (S3DIS) \cite{s3dis} dataset was collected using a Matterport scanner in 6 large indoor areas of 3 different buildings with a total of 272 rooms. \textcolor{red}{The dataset contains 13 semantic labels, including: ceiling (24.0\%), floor (23.8\%), wall (18.4\%), beam (1.8\%), column (1.5\%), window (1.3\%), door (4.0\%), table (3.8\%), chair (4.6\%), sofa (0.5\%), bookcase (4.5\%), board (0.7\%), and clutter (11.1\%).} In this study, each point is represented as a 6D vector (spatial coordinates ($xyz$) and color information ($rgb$)). We use a six-fold cross-validation to evaluate the proposed network. 

\subsubsection{Toronto3D}

The Toronto3D dataset \cite{Toronto3D} was obtained using a vehicle-mounted MLS system. The dataset was collected on Avenue Road in Toronto, Canada, covering approximately 1 km of the roadway. \textcolor{red}{The dataset was divided into four areas, each covering approximately 250 m, where each point is tagged with one of 8 semantic labels, including: road (55.6\%), road marking (2.4\%), nature (8.7\%), building (25.4\%), utility line (0.9\%), pole (1.2\%), car (5.2\%), and fence (0.6\%).} During the training, each point was represented as experiment on S3DIS. We strictly follow the authors' division of the training and test sets, which uses used Area 2 as the test set and the remaining three areas as the training set.

\subsubsection{SensatUrban}

The SensatUrban dataset \cite{SensatUrban} is a city-scale photogrammetric point cloud dataset containing nearly 3 billion points with detailed semantic annotation in 7.6 $km^2$. \textcolor{red}{Each point is tagged into one of 13 semantic labels, including: ground (20.3\%), vegetation (25.1\%), building (39.8\%), wall (1.0\%), bridge (0.1\%), parking (2.3\%), rail (0.02\%), car (1.7\%), footpath (2.0\%), bike (0.008\%), water (0.3\%), traffic road (6.1\%), and street furniture (1.2\%).} During the training, each point was represented as experiment on S3DIS. The dataset contained six regions distributed in two different cities (Birmingham and Cambridge) as online test sets. We strictly follow the authors' division of the training and test sets, which trains and tests on two different cities simultaneously.

\subsection{Semantic Segmentation Results}

\subsubsection{S3DIS}

Table \uppercase\expandafter{\romannumeral1} shows the quantitative results of our network compared with other state-of-the-art networks on the S3DIS dataset. Clearly, the proposed LACV-Net outperforms other state-of-the-art networks in terms of OA (89.7\%) and mIoU (72.7\%), which are 1.7\% and 2.7\% relative improvement over RandLA-Net \cite{RandLA-Net}, respectively. Our LACV-Net achieves significant performance on five out of thirteen classes: ceiling, beam, window, table, and board. We further conducted a more detailed comparison experiment on each area in the S3DIS dataset \cite{s3dis}. As shown in the Table \uppercase\expandafter{\romannumeral2}, we compare the results of the proposed LACV-Net with the selected state-of-the-art models \cite{RandLA-Net,SCF-Net,BAAF-Net} on S3DIS for each Area. The results of the comparison are taken from their papers. From the results it can be seen that the proposed LACV-Net achieved the best results on Area 1, 3, 4, 5, 6 and six-fold among the selected models. Therefore, we fully validate the effectiveness of our LACV-Net on the S3DIS dataset.

To intuitively compare the different methods, we visually compares the semantic segmentation results achieved by RandLA-Net \cite{RandLA-Net}, BAF-LAC \cite{BAF-LAC} and our LACV-Net on the S3DIS dataset as shown in Fig.5. The red box shows the part of RandLA-Net and BAF-LAC where the segmentation is wrong or the boundary segmentation is not significant. From the result, it is visually clear that LACV-Net segmentation of boards, columns, doors, and bookcases is superior to RandLA-net and BAF-LAC. In addition, the proposed LACV-Net can segment the boundaries of objects more smoothly and accurately. This is mainly because the proposed LAFA module learns adaptive weight by neighbor feature differences, whereas the proposed aggregated loss function further optimizes the adaptive weight so that the points near the boundaries are closer to their neighbors with similar features. Meanwhile, the color difference between the boundaries of different objects is evident, and the LAFA module encoding local color information can capture this color difference well.

\begin{table*}[ht]
\centering
\caption{Quantitative results of different approaches on the S3DIS dataset (six-fold cross-validation).}
\resizebox{\textwidth}{!}{%
\begin{tabular}{cccccccccccccccc}
\bottomrule
\multirow{2}*{Method}& \multirow{2}*{OA}& \multirow{2}*{mIoU}& \multirow{2}*{ceil.}& \multirow{2}*{floor}& \multirow{2}*{wall}& \multirow{2}*{beam}& \multirow{2}*{col.}& \multirow{2}*{wind.}& \multirow{2}*{door}& \multirow{2}*{table}& \multirow{2}*{chair}& \multirow{2}*{sofa}& \multirow{2}*{book.}& \multirow{2}*{board}& \multirow{2}*{clut.}\\
&&&&&&&&&&&&&&\\
\hline
PointNet \cite{PointNet} & 78.6 & 47.6 & 88.0 & 88.7 & 69.3 & 42.4 & 23.1 & 47.5 & 51.6 & 54.1 & 42.0 & 9.6 & 38.2 & 29.4 & 35.2 \\
SPGraph \cite{SPG} & 86.4 & 62.1 & 89.9 & 95.1 & 76.4 & 62.8 & 47.1 & 55.3 & 68.4 & 73.5 & 69.2 & 63.2 & 45.9 & 8.7 & 52.9 \\
AD-SAGC \cite{AD-SAGC} & 87.5 & 60.1 & 93.3 & 95.4 & 78.3 & 43.7 & 27.6 & 50.3 & 68.1 & 69.2 & 71.2 & 30.6 & 57.6 & 41.0 & 54.6 \\
PointWeb \cite{PointWeb} & 87.3 & 66.7 & 93.5 & 94.2 & 80.8 & 52.4 & 41.3 & 64.9 & 68.1 & 71.4 & 67.1 & 50.3 & 62.7 & 62.2 & 58.5 \\
ShellNet \cite{ShellNet} & 87.1 & 66.8 & 90.2 & 93.6 & 79.9 & 60.4 & 44.1 & 64.9 & 52.9 & 71.6 & 84.7 & 53.8 & 64.6 & 48.6 & 59.4 \\
KPConv \cite{KPConv} & -  & 70.6 & 93.6 & 92.4 & \textbf{83.1} & 63.9 & 54.3 & 66.1 & 76.6 & 57.8 & 64.0 & \textbf{69.3} & \textbf{74.9} & 61.3 & 60.3 \\
RandLA-Net \cite{RandLA-Net}  & 88.0 & 70.0 & 93.1 & 96.1 & 80.6 & 62.4 & 48.0 & 64.4 & 69.4 & 69.4 & 76.4 & 60.0 & 64.2 & 65.9 & 60.1 \\
BAAF-Net \cite{BAAF-Net} & 88.9 & 72.2 & 93.3 & \textbf{96.8} & 81.6 & 61.9 & 49.5 & 65.4 & 73.3 & 72.0 & \textbf{83.7} & 67.5 & 64.3 & 67.0 & \textbf{62.4} \\
BAF-LAC \cite{BAF-LAC} & 88.2 & 71.7 & 92.5 & 95.9 & 81.3 & 63.2 & \textbf{57.8} & 63.0 & \textbf{79.9} & 70.3 & 74.6 & 60.6 & 67.2 & 65.3 & 60.4 \\
SCF-Net \cite{SCF-Net} & 88.4 & 71.6 & 93.3 & 96.4 & 80.9 & 64.9 & 47.4 & 64.5 & 70.1 & 71.4 & 81.6 & 67.2 & 64.4 & 67.5 & 60.9 \\
LACV-Net(Ours) & \textbf{89.7} & \textbf{72.7} & \textbf{94.5} & 96.7 & 82.1 & \textbf{65.2} & 48.6 & \textbf{69.3} & 71.2 & \textbf{72.7} & 78.1 & 67.3 & 67.2 & \textbf{70.9} & 61.6 \\
\bottomrule
\end{tabular}%
}
\end{table*}

\begin{table}[]
\centering
\caption{Detailed semantic segmentation results on the S3DIS dataset.}
\resizebox{0.48\textwidth}{!}{%
\begin{tabular}{ccccccc|c}
\bottomrule
{Method} &  {Area 1} & {Area 2} & {Area 3} & {Area 4} & {Area 5} & {Area 6} &6-fold\\
\hline
RandLA-Net \cite{RandLA-Net} & 74.5 & 56.2 & 78.1 & 61.7 & 62.5 & 80.3 &70.0\\
SCF-Net \cite{SCF-Net} & 75.1 & \textbf{59.7} & 78.4 & 60.2 & 63.4 & 80.2 &71.6\\
BAAF-Net \cite{BAAF-Net} & 76.3 & 57.8 & 80.0 & 64.3 & 65.4 & 81.8 &72.2\\
LACV-Net (Ours) & \textbf{77.0} & 57.6 & \textbf{80.4} & \textbf{64.4} & \textbf{65.9} & \textbf{82.2} & \textbf{72.7}\\
\bottomrule
\end{tabular}%
}
\vspace{-1.0em}
\end{table}

\begin{figure*}[ht]
\centering
\includegraphics[width=1.0\textwidth]{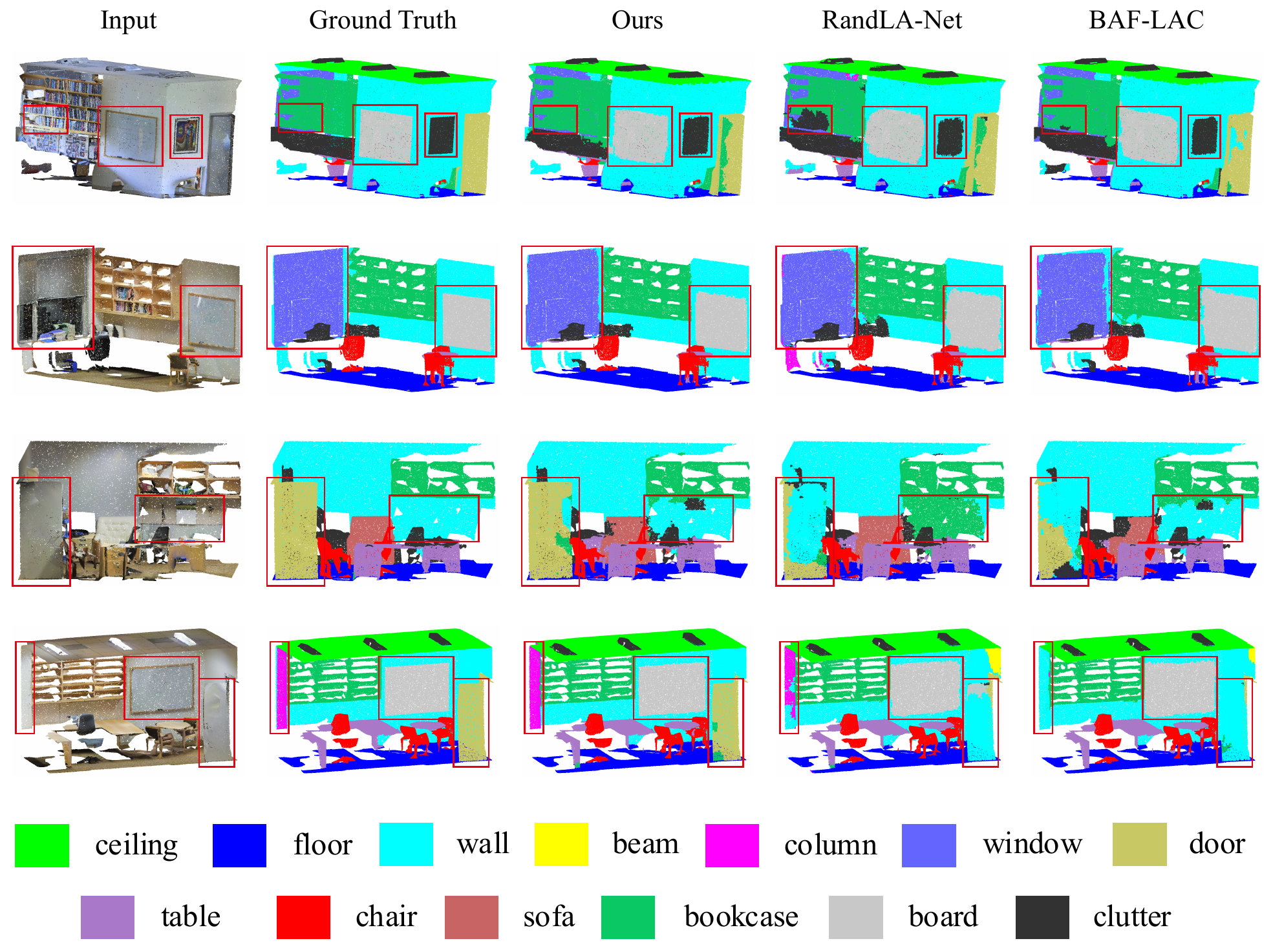}
\caption{Visual comparison of semantic segmentation results on S3DIS dataset.}
\label{}
\vspace{-1.0em}
\end{figure*}

\subsubsection{Toronto3D}

Table \uppercase\expandafter{\romannumeral3} shows the quantitative results of our network compared with other state-of-the-art networks on the Toronto3D dataset. \textcolor{red}{Since some of the methods without use $rgb$ in the Toronto3D dataset, for the sake of fairness, and to verify the performance of the model on point cloud data without spectral information, we implement two groups of experiments, the first group we do not use $rgb$ information and compared with the method without using $rgb$, and the second group we use $rgb$ information and only compared with the method using $rgb$.} The results show that when using $rgb$ information as initial input, the proposed LACV-Net achieves excellent performance in terms of OA (97.4\%) and mIoU (82.7\%), which are 3.0\% and 0.9\% relative improvement over RandLA-Net \cite{RandLA-Net}, respectively. The proposed LACV-Net achieves superior performance on 
three out of the eight classes: road, natural and fence. We also experiment without using $rgb$ information as initial input. In this case, the proposed LACV-Net achieves excellent performance in terms of OA (95.8\%) and mIoU (78.5\%), which are 2.8\% and 0.8\% relative improvement over RandLA-Net \cite{RandLA-Net}, respectively.

To intuitively compare the different methods, Fig.6 qualitatively shows the visual comparison achieved by RandLA-Net \cite{RandLA-Net}, BAF-LAC \cite{BAF-LAC} and our LACV-Net on the Toronto3D dataset using $rgb$ as initial input. The red boxes show the part of RandLA-Net and BAF-LAC where the segmentation is wrong or the boundary segmentation is not significant. From the results we can observe that LACV-Net segmentation of car, natural and fence is significantly superior to that of RandLA-Net and BAF-LAC. In particular, for the semantic segmentation of the crosswalk, the proposed LACV-Net is more accurate and clear in the boundaries, comparing to the excellent Rand-Net and BAF-LAC. This is mainly because there is almost no difference between the geometric coordinates of road markers and roads, and compared with RandLA-Net and BAF-LAC which only encode spatial geometric information, the proposed local information encoding unit explicitly encodes color information, allowing the network to fully exploit local color differences. \textcolor{red}{In addition, as shown in Fig.7, we visually compared the segmentation result with or without using spectral information $rgb$ on the Toronto3D dataset. From the visualization results, it can be observed that for the class of road markings, the results differ significantly, and for the other classes, the difference of results is less. }

\begin{table*}[]
\centering
\caption{Quantitative results of different approaches on Toronto3D dataset (Area 2).}
\resizebox{\textwidth}{!}{%
\begin{tabular}{cccccccccccccccc}
\bottomrule
\multirow{2}*{RGB}&\multirow{2}*{Method}& \multirow{2}*{OA}& \multirow{2}*{mIoU}& \multirow{2}*{Road}& \multirow{2}*{Road mrk.}& \multirow{2}*{Nature}& \multirow{2}*{Buil.}& \multirow{2}*{Util. line}& \multirow{2}*{Pole}& \multirow{2}*{Car}& \multirow{2}*{Fence}\\
&&&&&&&&&&\\
\hline
\multirow{8}*{No}&PointNet++ \cite{PointNet++} & 92.6 & 59.5 & 92.9 & 0.0 & 86.1 & 82.2 & 60.9 & 62.8 & 76.4 & 14.4 \\
~&DGCNN \cite{DGCNN} & 94.2 & 61.7 & 93.9 & 0.0 & 91.3 & 80.4 & 62.4 & 62.3 & 88.3 & 15.8 \\
~&MS-PCNN \cite{MS-PCNN} & 90.0 & 65.9 & 93.8 & 3.8 & 93.5 & 82.6 & 67.8 & 71.9 & 91.1 & 22.5 \\
~&KPConv \cite{KPConv} & 95.4 & 69.1 & 94.6 & 0.1 & 96.1 & 91.5 & 87.7 & \textbf{81.6} & 85.7 & 15.7 \\
~&TGNet \cite{TGNet} & 94.1 & 61.3 & 93.5 & 0.0 & 90.8 & 81.6 & 65.3 & 62.9 & 88.7 & 7.9 \\
~&MS-TGNet \cite{Toronto3D} & 95.7 & 70.5 & 94.4 & 17.2 & 95.7 & 88.8 & 76.0 & 73.9 & \textbf{94.2} & 23.6 \\
~&RandLA-Net \cite{RandLA-Net} & 93.0 & 77.7 & 94.6 & 42.6 & \textbf{96.9} & \textbf{93.0} & 86.5 & 78.1 & 92.9 & 37.1\\
~&LACV-Net (Ours) & \textbf{95.8} & \textbf{78.5} & \textbf{94.8} & \textbf{42.7} & 96.7 & 91.4 & \textbf{88.2} & 79.6 & 93.9 & \textbf{40.6}\\
\hline
\multirow{5}*{Yes}&RandLA-Net \cite{RandLA-Net} & 94.4 & 81.8 & 96.7 & 64.2 & 96.9 & \textbf{94.2} & \textbf{88.0} & 77.8 & 93.4 & 42.9 \\
~&ResDLPS-Net \cite{ResDLPS-Net} & 96.5 & 80.3 & 95.8 & 59.8 & 96.1 & 90.9 & 86.8 & 79.9 & 89.4 & 43.3\\
~&BAF-LAC \cite{BAF-LAC} & 95.2 & 82.2 & 96.6 & 64.7 & 96.4 & 92.8 & 86.1 & \textbf{83.9} & \textbf{93.7} & 43.5\\
~&BAAF-Net \cite{BAAF-Net} & 94.2 & 81.2 & 96.8 & \textbf{67.3} & 96.8 & 92.2 & 86.8 & 82.3 & 93.1 & 34.0\\
~&LACV-Net (Ours) & \textbf{97.4} & \textbf{82.7} & \textbf{97.1} & 66.9 & \textbf{97.3} & 93.0 & 87.3 & 83.4 & 93.4 & \textbf{43.1}\\
\bottomrule
\end{tabular}%
}
\end{table*}

\begin{figure*}
\centering
\includegraphics[width=1.0\textwidth]{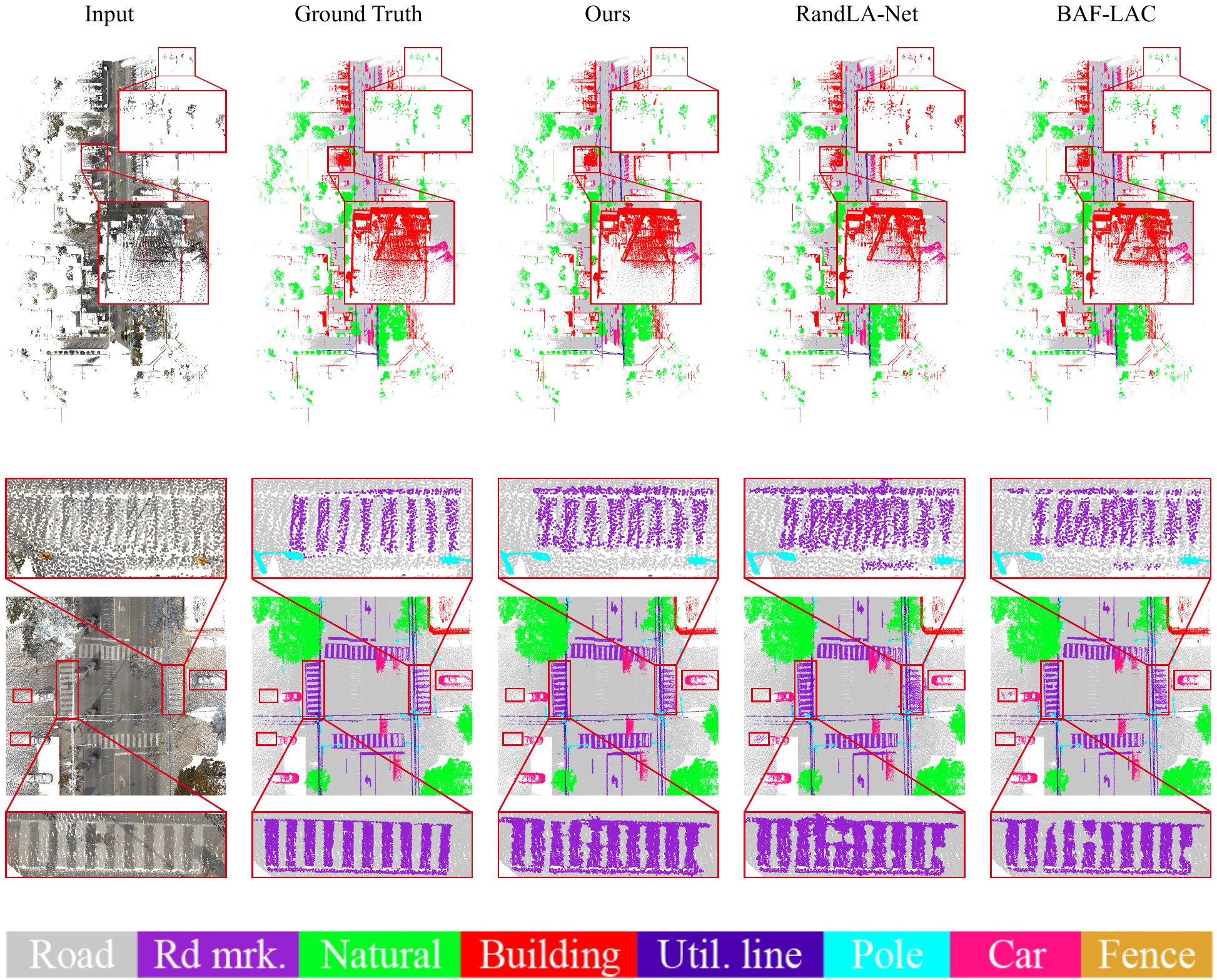}
\caption{Visual comparison of semantic segmentation results on Toronto3D dataset.}
\label{}
\end{figure*}

\begin{figure}
\centering
\includegraphics[width=0.48\textwidth]{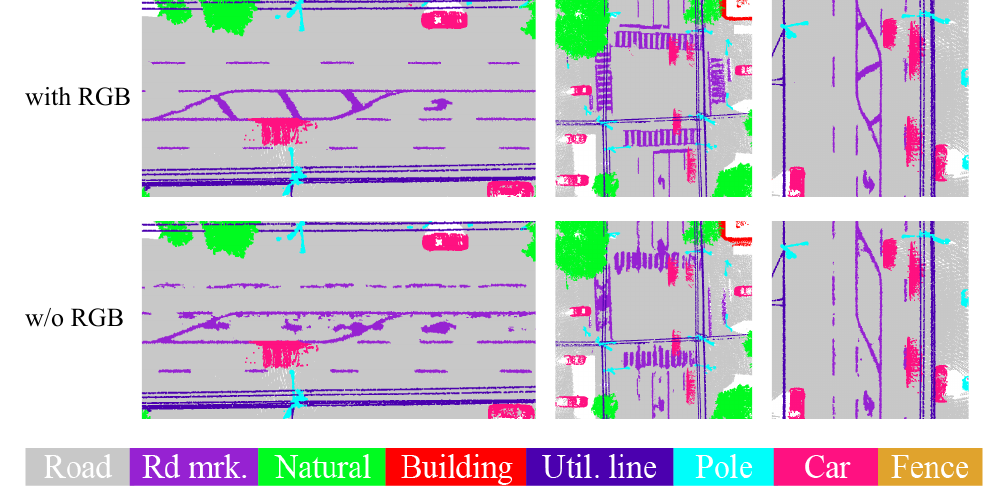}
\caption{Visual comparison of with or without $rgb$ as initial input on Toronto3D dataset.}
\label{}
\end{figure}

\subsubsection{SensatUrban}

Table \uppercase\expandafter{\romannumeral4} summarizes the quantitative results of our network compared with other state-of-the-art networks on the SensatUrban dataset. The results show that the proposed LACV-Net achieves excellent performance in terms of OA (93.2\%) and mIoU (61.3\%), which are 3.4\% and 8.4\% relative improvements over RandLA-Net, respectively. The proposed LACV-Net achieves superior performance on seven out of the thirteen classes: buildings, bridge, parking, rail, traffic roads, footpath and bikes. We notice that training and testing on the same city substantially improved the accuracy of the network \cite{SensatUrban}, but for fairness, we still train and test on two different cities simultaneously.

Fig.8 qualitatively shows the visual comparison achieved by RandLA-Net \cite{RandLA-Net}, BAF-LAC \cite{BAF-LAC} and our LACV-Net on the SensatUrban dataset. Since the ground truth of the online test sets are not publicly available, to better demonstrate the visualization, we show the results on the validation set. The red box shows the part of RandLA-Net and BAF-LAC where the segmentation is wrong or the boundary segmentation is not significant. It can be observed that compared to RandLA-Net and BAF-LAC, which barely segment the railway, the proposed LACV-Net can clearly segment the railway from the ground. Meanwhile, the proposed LACV-Net has a significant improvement in the performance of parking and traffic road. Considering the results of both the qualitative and quantitative analyses, the proposed LACV-Net can accurately classify the semantic labels of large-scale point cloud scenes. 

\begin{table*}[]
\centering
\caption{Quantitative results of different approaches on SenastUrban dataset.}
\resizebox{\textwidth}{!}{%
\begin{tabular}{cccccccccccccccc}
\bottomrule
\multirow{2}*{Method}& \multirow{2}*{OA}& \multirow{2}*{mIoU}& \multirow{2}*{ground}& \multirow{2}*{veg.}& \multirow{2}*{buil.}& \multirow{2}*{wall}& \multirow{2}*{bri.}& \multirow{2}*{park.}& \multirow{2}*{rail}& \multirow{2}*{traffic.}& \multirow{2}*{street.}& \multirow{2}*{car}& \multirow{2}*{foot.}& \multirow{2}*{bike}& \multirow{2}*{water}\\
&&&&&&&&&&&&&&\\
\hline
PointNet \cite{PointNet} & 80.8 & 23.7 & 67.9 & 89.5 & 80.1 & 0.0 & 0.0 & 3.9 & 0.0 & 31.6 & 0.0 & 35.1 & 0.0 & 0.0 & 0.0 \\
PointNet++ \cite{PointNet++} & 84.3 & 32.9 & 72.5 & 94.2 & 84.8 & 2.7 & 2.1 & 25.8 & 0.0 & 31.5 & 11.4 & 38.8 & 7.1 & 0.0 & 56.9 \\
TagentConv \cite{TagentConv} & 76.9 & 33.3 & 71.5 & 91.4 & 75.9 & 35.2 & 0.0 & 45.3 & 0.0 & 26.7 & 19.2 & 67.6 & 0.0 & 0.0 & 0.0 \\
SPGraph \cite{SPG} & 85.3 & 37.3 & 69.9 & 94.6 & 88.9 & 32.8 & 12.6 & 15.8 & 15.5 & 30.6 & 22.9 & 56.4 & 0.5 & 0.0 & 44.2 \\
SparseConv \cite{SparseConv} & 88.7 & 42.7 & 74.1 & 97.9 & 94.2 & 63.3 & 7.5 & 24.2 & 0.0 & 30.1 & 34.0 & 74.4 & 0.0 & 0.0 & 54.8 \\
KPConv \cite{KPConv} & 93.2 & 57.6 & 87.1 & \textbf{98.9} & 95.3 & \textbf{74.4} & 28.7 & 41.4 & 0.0 & 55.9 & \textbf{54.4} & \textbf{85.7} & 40.4 & 0.0 & \textbf{86.3} \\
RandLA-Net \cite{RandLA-Net}  & 89.8 & 52.7 & 80.1 & 98.1 & 91.6 & 48.9 & 40.6 & 51.6 & 0.0 & 56.7 & 33.2 & 80.1 & 32.6 & 0.0 & 71.3 \\
BAF-LAC \cite{BAF-LAC} & 91.5 & 54.1 & 84.4 & 98.4 & 94.1 & 57.2 & 27.6 & 42.5 & 15.0 & 51.6 & 39.5 & 78.1 & 40.1 & 0.0 & 75.2 \\
BAAF-Net \cite{BAAF-Net} & 92.0 & 57.3 & 84.2 & 98.3 & 94.0 & 55.2 & 48.9 & 57.7 & 20.0 & 57.3 & 39.3 & 79.3 & 40.7 & 0.0 & 70.1 \\
LACV-Net(Ours) & \textbf{93.2} & \textbf{61.3} & \textbf{85.5} & 98.4 & \textbf{95.6} & 61.9 & \textbf{58.6} & \textbf{64.0} & \textbf{28.5} & \textbf{62.8} & 45.4 & 81.9 & \textbf{42.4} & \textbf{4.8} & 67.7 \\
\bottomrule
\end{tabular}%
}
\end{table*}

\begin{figure*}
\centering
\includegraphics[width=1.0\textwidth]{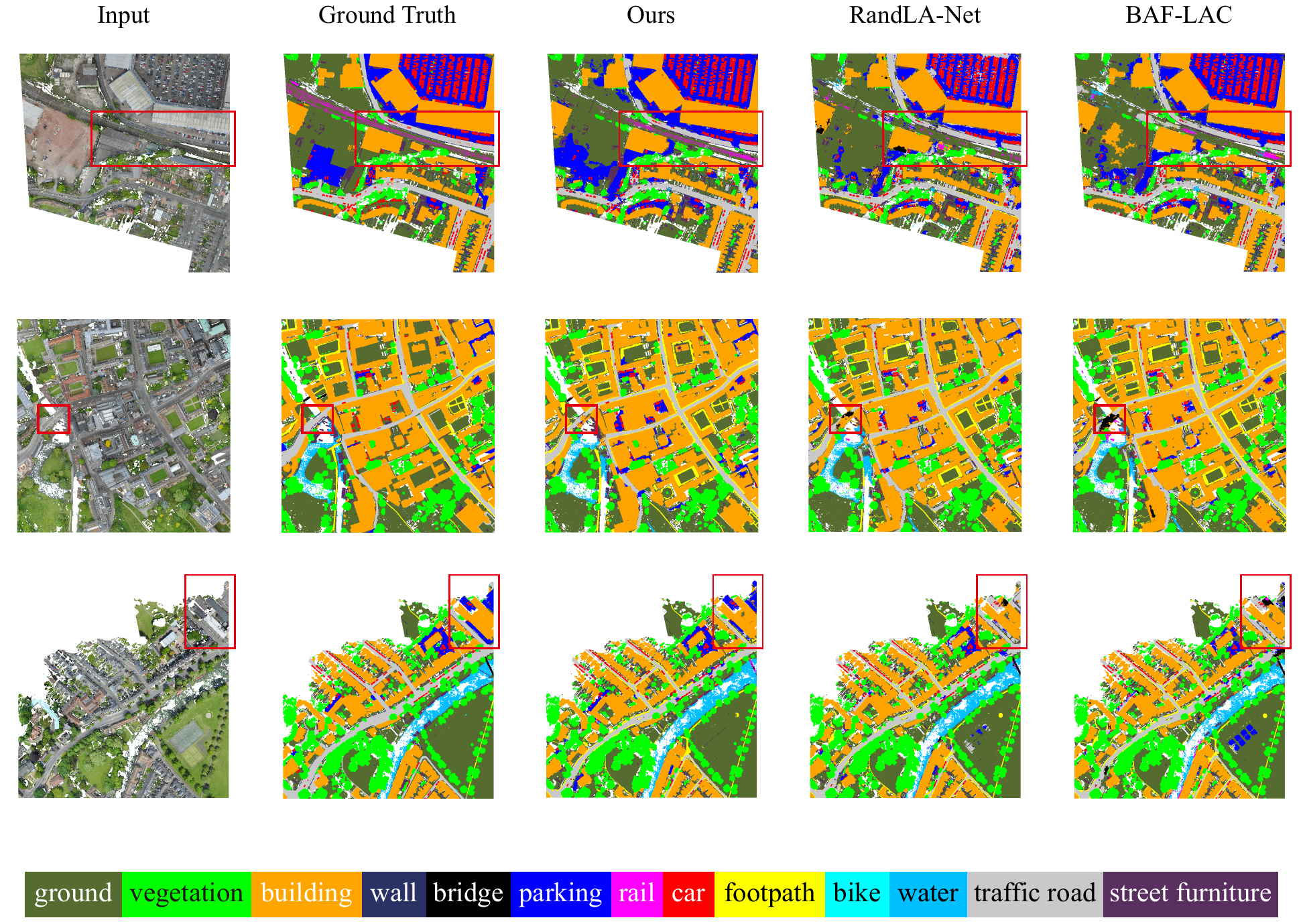}
\caption{Visual comparison of semantic segmentation results on SensatUrban dataset.}
\label{}
\vspace{-1.0em}
\end{figure*}

\subsection{Ablation Studies}

In this section, to quantitatively or qualitatively evaluate the effectiveness of the proposed modules, and the units within each module, we conduct ablation studies for efficiency of LACV-Net framework, local information encoding unit, adaptive augmentation unit, comprehensive VLAD module, and aggregation loss function. All ablation networks are trained on S3DIS \cite{s3dis} area 1-4 and 6, and tested on S3DIS area 5.

\subsubsection{Ablation of LACV-Net framework} The proposed LACV-Net is mainly composed of LAFA module and C-VLAD module, while the LAFA module contains local information encoding unit and adaptive augmentation unit. To demonstrate the effectiveness of each component in the proposed LACV-Net, we conduct the following three ablation studies to demonstrate the effectiveness of each component in the LACV-Net. (1) Remove adaptive augmentation unit. The adaptive augmentation unit can adaptively learn the similarity of centroids and neighboring points to avoid the problem of neighbor perception ambiguity. After removing adaptive augmentation unit, we use a mean-pooling, which only serves to compress the feature dimensions output from the previous local information encoding unit. (2) Remove local information encoding unit. local information encoding unit can efficiently encode the local spatial, color and semantic information to efficiently augment the local context. After removing local information encoding unit, we repeat the semantic features $k$ times, and directly use a $mlp$ to align the dimensions which is required for the next adaptive augmentation unit ($\mathbb{R}^{N\times 1\times C_{in}} \rightarrow \mathbb{R}^{N\times k\times C_{out}}$). (3) Remove the C-VLAD module. The C-VLAD module can fuse local features with multi-layer, multi-scale and multi-resolution to represent a comprehensive global description vector. After removing C-VLAD module, we directly output the last layer of the encoder to the decoder.

Table \uppercase\expandafter{\romannumeral5} shows the results of the ablation studies. From the results, it is clear that: (1) The removing of the local information encoding unit shows the greatest impact on performance, demonstrating that this module is necessary to effectively learn the local context. (2) The removing of adaptive augmentation unit shows the next greatest impact on performance. This is highlighted in Fig 3-(b), which shows how assigning the points near the boundary to the correct label by adaptive weight. (3) The removing of C-VLAD module diminishes performance by not being able to capture the global information distributed over a large spatial scale. From this ablation study, we verify the effectiveness of the proposed component, and it can be observed how they complement each other to achieve state-of-the-art performance. The detailed ablation studies for the effectiveness of the units within each module are introduced in the follow-up.

\begin{table}[]
\centering
\setlength{\tabcolsep}{2mm}
\caption{Ablation studies of LACV-Net framework. $LIE$: local information encoding, $AA$: adaptive augmentation.}
\resizebox{0.45\textwidth}{!}{%
\begin{tabular}{ccccc}
\bottomrule
 & $LIE$ unit & $AA$ unit & C-VLAD module & mIoU \\
\hline
A1 & & \checkmark & \checkmark & 57.2 \\
A2 & \checkmark & & \checkmark & 62.4 \\
A3 & \checkmark & \checkmark & & 64.6 \\
\hline
\textbf{A4} & \checkmark & \checkmark & \checkmark & \textbf{65.9} \\
\bottomrule
\end{tabular}%
}
\vspace{-1.0em}
\end{table}

\subsubsection{Ablation of Local Information Encoding unit} As designed in Section 3.2, the proposed local information encoding unit is designed to make full use of the local neighbor information, which fuses all the neigboring spatial/color/semantic information to the centroid, based on the following equation: 

{\setlength\abovedisplayskip{1.5pt}
\begin{align}
\vspace{-1.0em}
\triangle \ell_i=\{\ell(p_i)\oplus\ell(c_i)\oplus\ell(f_i)\}
\end{align}
}

In this section, we investigate the performances of spatial information encoding, color information encoding, and semantic information encoding on the LFE module. Spatial coordinates $xyz$ and color information $rgb$ are used as initial input to each ablation network. We conduct the following ablative experiments. (1) Only use the semantic information encoding. (2) Use the semantic information encoding with spatial information encoding. (3) Use the semantic information encoding with color information encoding. (4) Use the spatial information encoding with color information encoding, in this case, semantic features of the previous layer are concatenated with this encoding. (5) Use all the spatial, color and semantic information encodings.

Table \uppercase\expandafter{\romannumeral6} shows the results of the ablation studies. From the results, it is clear that: (1) The spatial information encoding shows the most dominant impact on performance, mainly because of its ability to preserve local geometric structure. (2) The color information encoding shows secondary impact on performance. It is shown that even if the color information is acquired with some noise, the encoding of the color information can still learn local color differences. (3) The impact of semantic information encoding on performance is comparable to that of color information encoding. In summary, the validity of spatial/color/semantic encoding are demonstrated.

\begin{table}[]
\centering
\setlength{\tabcolsep}{2mm}
\caption{Ablation studies of local information encoding unit. $f$: local semantic information, $xyz$: local spatial information, $rgb$: local color information.}
\begin{tabular}{ccc}
\bottomrule
 & Encoding & mIoU \\
\hline
B1 & only $f$ &  58.7 \\
B2 & $xyz$ + $f$ & 64.0 \\
B3 & $rgb$ + $f$ & 60.5 \\
B4 & $xyz$ + $rgb$ & 64.4 \\
\hline
\textbf{B5} & $xyz$ + $rgb$ + $f$ & \textbf{65.9} \\
\bottomrule
\end{tabular}%
\vspace{-1.0em}
\end{table}

\subsubsection{Ablation of Adaptive Augmentation unit} The proposed adaptive augmentation unit is designed to solve the problem of ambiguous neighbor perception by calculating the adaptive weight via similarity between the local information of neighboring points and the centroid. We combine the max and sum pooling operations to enhance the network capture of the entire local neighbor and the local significant features. The adaptive augmentation unit is based on the following equation: 

{\setlength\abovedisplayskip{1.5pt}
\begin{align}
\vspace{-1.0em}
F_i^c=mlp\{(\sum_{k=1}^{K}{(W_i^c\triangle\ell_i^c)}_k) \oplus (max{(\triangle\ell_i^c)}_k)\}
\end{align}
}

In this section, we investigate the performances of the designed adaptive weight and different pooling operations. Since the computation of the designed aggregated loss function requires adaptive weight, in this ablation study, aggregated loss function is removed to fairly compare the performance of adaptive weight. We conduct the following ablative experiments. (1-2) Only use the max/sum pooling operation without using adaptive weight. (3) Use max and sum pooling operations without using adaptive weight. (4-5) Use the max/sum pooling operation and using adaptive weight. (6) Use max and sum pooling operations with using adaptive weight.

Table \uppercase\expandafter{\romannumeral7} shows the results of the ablation studies. From the results, it is clear that: (1) Compared to the network without adaptive weight, the accuracy of that with adaptive weight has a significant improvement, regardless of the pooling operation used. (2) The performance of max pooling and sum pooling operation is comparable, and both are inferior to combined max and sum pooling operations. In summary, the validity of adaptive weight and pooling operations are demonstrated.

In addition, we qualitatively analyzed the role of adaptive weight on semantic segmentation. As shown in the Fig.9, the red box shows the part of network without adaptive weight where the segmentation is wrong or the boundary segmentation is not significant. It can be observed that for large planes objects, such as walls and floors, the feature maps are clear with or without adaptive weight. But for uneven objects, such as bookcases, tables, chairs, and clutter, the feature maps of the network without adaptive weight ambiguous in local perception, while with adaptive weight are smoother and the boundaries are more accurate and clear. This is mainly because the adaptive weight Learns the influence of neighbor points on the centroid to make the features of centroids closer to their similar neighbors, which can solve the problem of ambiguous local perception and fully utilize the local information.

\begin{table}[]
\centering
\vspace{0.5em}
\setlength{\tabcolsep}{2mm}
\caption{Ablation studies of adaptive augmentation unit.}
\begin{tabular}{cccc}
\bottomrule
 & Weight & Pooling & mIoU \\
\hline
C1 & none & max & 62.8 \\
C2 & none & sum & 63.0 \\
C3 & none & max + sum &  63.6\\
C4 & adaptive weight & max &  64.1 \\
C5 & adaptive weight & sum &  64.0 \\
\hline
\textbf{C6} & adaptive weight & max + sum &  \textbf{64.7} \\
\bottomrule
\end{tabular}%
\vspace{-1.0em}
\end{table}

\begin{figure*}
\centering
\includegraphics[width=1.0\textwidth]{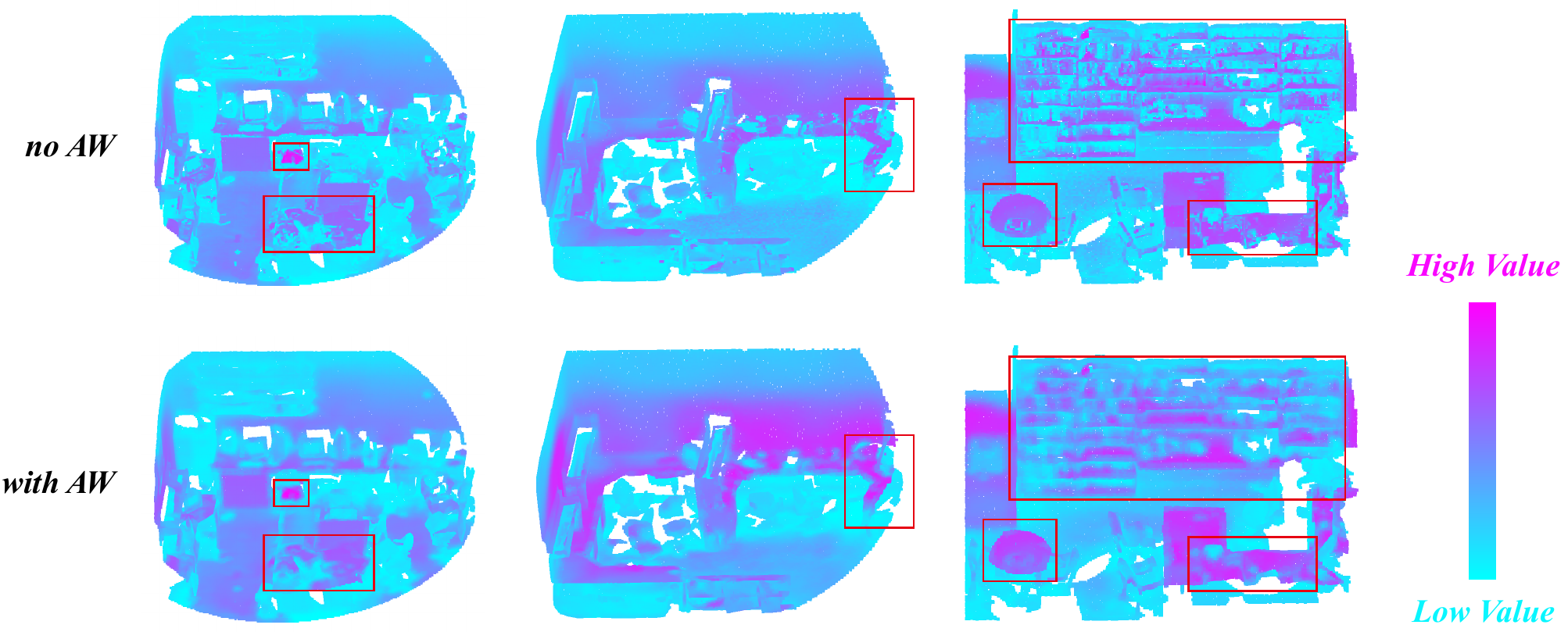}
\caption{Visual comparison of feature map on S3DIS dataset. $AW$: adaptive weight.}
\label{}
\vspace{-1.0em}
\end{figure*}

\subsubsection{Ablation of Comprehensive VLAD module} The proposed C-VLAD module is designed to capture global information from 3D point clouds by fusing local features with multi-layer, multi-scale and multi-resolution and solve the problem of global feature representation insignificant. We input the output of the encoding layer into the C-VLAD module, which is designed based on original VLAD module \cite{VLAD}. In this section, we investigate the compared performances of the designed C-VLAD module and other widely used obtaining global features operations. We conduct the following ablative experiments. (1) Removing C-VLAD module as the baseline network. (2-4) Replacing C-VLAD module with original VLAD/max-pooling/mean-pooling operation. (5) Using the proposed C-VLAD module.

Table \uppercase\expandafter{\romannumeral8} shows the results of the ablation studies. From the results, it is clear that the original VLAD, max-pooling and mean-pooling are performed comparably, and the proposed C-VLAD module achieves the best performance among them. It is mainly because in the process of down-sampling, the total number of points is reduced and the perceptual field of the LAFA module is relatively dilated to capture a wider range of local neighbor information, which enhances the perception of the global. Meanwhile, global max or mean pooling loses substantial detailed features, resulting in insufficient global feature mappings for fine-grained semantic segmentation. Whereas the original VLAD uses only the output of the last encoding layer as input, whose output features become implicit and abstract after several down-sampling operations. Meanwhile, the proposed C-VLAD module fusing local features with multi-layer, multi-scale and multi-resolution to capture global features can alleviate the above problems. In summary, the validity of C-VLAD module is demonstrated.

\begin{table}[]
\centering
\setlength{\tabcolsep}{2mm}
\caption{Ablation studies of Comprehensive-VLAD module.}
\begin{tabular}{clc}
\bottomrule
 & & mIoU \\
\hline
D1 & Remove C-VLAD module & 64.6 \\
D2 & Replace with original VLAD & 65.2 \\
D3 & Replace with max-pooling & 65.1 \\
D4 & Replace with mean-pooling & 64.9 \\
\hline
\textbf{D5} & Comprehensive VLAD (Ours) & \textbf{65.9} \\
\bottomrule
\end{tabular}%
\vspace{-1.0em}
\end{table}

\subsubsection{Ablation of Aggregated Loss} The aggregated loss function is designed based on widely used weighted cross-entropy loss function to further optimize the adaptive weight that better represent the influence of the neighboring points on the centroids, such that the points in the vicinity of the boundary can be correctly classified, to improve the performance of the network. The proposed aggregated loss function consists of the proposed constraint loss function and the widely used weighted cross-entropy loss function. In this section, we compare the network training with aggregated loss function and the weighted cross-entropy loss function only. We conduct the following ablative experiments. (1) Only using the weighted cross-entropy loss function. (2) Using the aggregated loss function.

As shown in Fig.10-(a), it can be clearly seen that although the loss value of the aggregated loss function is initially higher than that of the weighted cross-entropy loss function, the convergence speed of the aggregated loss function is significantly better than that of the weighted cross-entropy loss function. Meanwhile, as shown in the Fig.9-(b), the network using the aggregated loss function shows a better fitting performance on the training set than that of using the weighted cross-entropy loss function only. In addition, as shown in Table \uppercase\expandafter{\romannumeral9}, compared to the network only using the weighted cross-entropy loss function, the accuracy of that using the aggregated loss function has a significant improvement. In summary, the validity of the aggregated loss function is demonstrated. 

\begin{figure}
\centering
\includegraphics[width=0.48\textwidth]{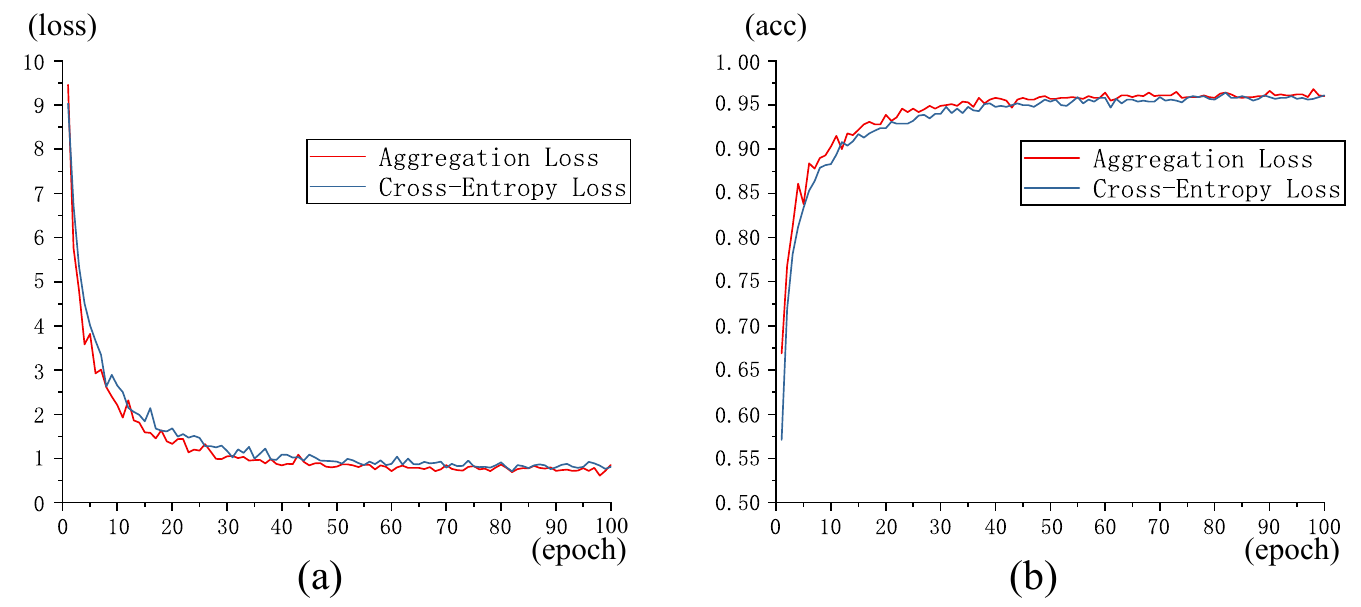}
\caption{Losses and performance on training set with the proposed aggregation loss and weighted cross-entropy loss.\\
(a) losses. (b) overall accuracy.}
\label{}
\vspace{-0.5em}
\end{figure}

\begin{table}[]
\centering
\setlength{\tabcolsep}{2mm}
\caption{Comparison of different loss functions.}
\begin{tabular}{llc}
\bottomrule
 & & mIoU \\
\hline
E1 & weighted cross-entropy loss only & 64.7 \\
\textbf{E2} & aggregated loss (Ours) & \textbf{65.9} \\
\bottomrule
\end{tabular}%
\vspace{-1.0em}
\end{table}

\begin{table}[]
\vspace{0.5em}
\centering
\caption{Computation efficiency and parameters on S3DIS dataset.}
\setlength{\tabcolsep}{1mm}{
\begin{tabular}{ccccccc}
\bottomrule
{} & {mIoU} & & {Per-batch time} & & {Total inference time} \\
 & (\%) & & (ms) & & (s) \\
\hline
RandLA-Net \cite{RandLA-Net} & 70.0 && 426.3 && 262.3  \\
BAAF-Net \cite{BAAF-Net} & 72.2 && 670.5 && 245.3  \\
BAF-LAC \cite{BAF-LAC} & 71.7 && 584.5 && 266.4  \\
LACV-Net & 72.7 && 645.0 && 262.8  \\
\bottomrule
\end{tabular}}%
\vspace{0.5em}
\end{table}

\begin{table}[]
\centering
\caption{Quantitative results of computation confidence.}
\setlength{\tabcolsep}{1mm}{
\begin{tabular}{ccccccc}
\bottomrule
\multirow{2}*{S3DIS (Area 5)} && \multirow{2}*{Toronto3D (Area 2)} && \multirow{2}*{SensatUrban (Val set)} \\
&&&&\\
\hline
66.0 $\pm$ \textcolor{red}{0.4} && 82.4 $\pm$ \textcolor{red}{0.3} && 52.5 $\pm$ \textcolor{red}{1.0} \\
\bottomrule
\end{tabular}}%
\end{table}

\subsection{Efficiency analysis}

We systematically evaluated the computational efficiency and number of parameters of our model compare with other excellent large-scale point cloud semantic segmentation models on S3DIS. In Table \uppercase\expandafter{\romannumeral10}, we selected the per-batch training time and total inference time as computational efficiency metrics, while mIoU as performance metrics. RandLA-Net \cite{RandLA-Net} has the shortest per-batch training time among all models, but its performance metrics are also the lowest. BAAF-Net \cite{BAAF-Net} has the longest per-batch training time because of its use of expensive farthest point sampling, and BAF-LAC \cite{BAF-LAC} has the longest inference time. The proposed LACV-Net outperforms these superior networks in terms of accuracy, and is comparable to them in terms of computational efficiency metrics. Therefore, the proposed LACV-Net can be adapted for large-scale point cloud data.

{\color{red}\subsection{Confidence interval analysis}}

\textcolor{red}{We evaluated the computation confidence of the proposed LACV-Net on the used three datasets. Note that, since SensatUrban's test sets are online, the validation sets of SensatUrban were used in this study. We used the same experimental setting, and reported the mean$\pm$std in three random runs. As shown in Table \uppercase\expandafter{\romannumeral11}, our method can be easily reproduced, and the standard deviation of three random runs is less than 1.0. Since SensatUrban is a city-scale dataset with a considerable scene size and unbalanced classes distribution, it has a larger variance compared to S3DIS and Toronto3D.}

\section{Conclusion}

In this study, an end-to-end network LACV-Net is proposed for large-scale point cloud scene semantic segmentation. The main contributions of this study are as follows: First, a local adaptive feature augmentation module (LAFA) is proposed to efficiently learn the local contextual. We experimentally demonstrate that the internal local information encoding unit can augment the local context, while the internal adaptive augmentation unit can effectively alleviate the problem of local perceptual ambiguity. Subsequently, a comprehensive VLAD module (C-VLAD) is proposed to fuse comprehensive global knowledge. Compared to the widely used global maximum, average pooling and traditional VLAD, the proposed C-VLAD demonstrates superior performance in global feature representation. Finally, an aggregation loss function is proposed to effectively constrain the adaptive weight from the LAFA module. The experimental results demonstrate that the designed aggregation loss function improves accuracy and accelerates the convergence of the network. We achieve excellent performance on three benchmark datasets comparing with state-of-the-art networks, including indoor dataset S3DIS \cite{s3dis} and outdoor dataset Toronto3D \cite{Toronto3D} and SensatUrban\cite{SensatUrban}. 

\section{Acknowledgment}
This study is funded by National Key Research and Development Program of China [2021YFB2600300, 2021YFB2600302].

\bibliographystyle{IEEEtran}
\bibliography{egbib}

\vspace{-33pt}
\begin{IEEEbiography}[{\includegraphics[width=1in,height=1.25in,clip,keepaspectratio]{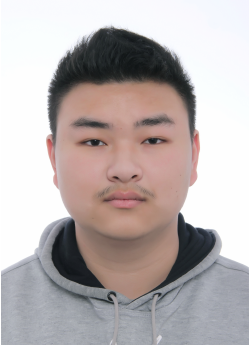}}]{Ziyin Zeng}
received the bachelor’s degree in Safety Engineering from the Zhongnan University of Economics and Law, Wuhan, China, in 2020. He is currently pursuing the master’s degree with the China University of Geosciences, Wuhan, China. His research interest includes LiDAR Remote Sensing, computer vision and point cloud processing, such as point cloud semantic segmentation, 3D reconstruction from multi-source remote sensing data and graph-structured data applications.
\end{IEEEbiography}

\vspace{-33pt}
\begin{IEEEbiography}[{\includegraphics[width=1in,height=1.25in,clip,keepaspectratio]{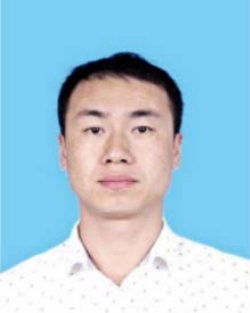}}]{Yongyang Xu}
received the B.S. degrees in computer science and technology from China University of Geosciences (Wuhan), in 2014, and the Ph.D. degree in geographic information engineering from China University of Geosciences (Wuhan), in 2019. He is currently an associate professor in School of Geography and Information Engineering, China University of Geosciences (Wuhan). His main interests include deep learning, vector data rendering and processing.
\end{IEEEbiography}

\vspace{-33pt}
\begin{IEEEbiography}[{\includegraphics[width=1in,height=1.25in,clip,keepaspectratio]{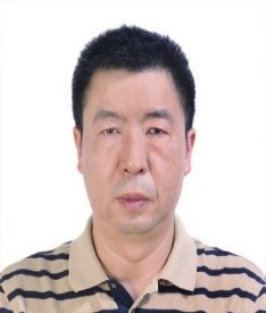}}]{Zhong Xie}
received the B.S., M.S., and Ph.D. degrees in cartography and geographic information engineering from the China University of Geosciences, Wuhan, China, in 1990, 1998, and 2002, respectively. He is currently a Professor with the School of Geography and Information Engineering, China University of Geosciences. His research interests include deep learning, 3-D rebuilding and spatial analysis, image processing.
\end{IEEEbiography}

\vspace{-33pt}
\begin{IEEEbiography}[{\includegraphics[width=1in,height=1.25in,clip,keepaspectratio]{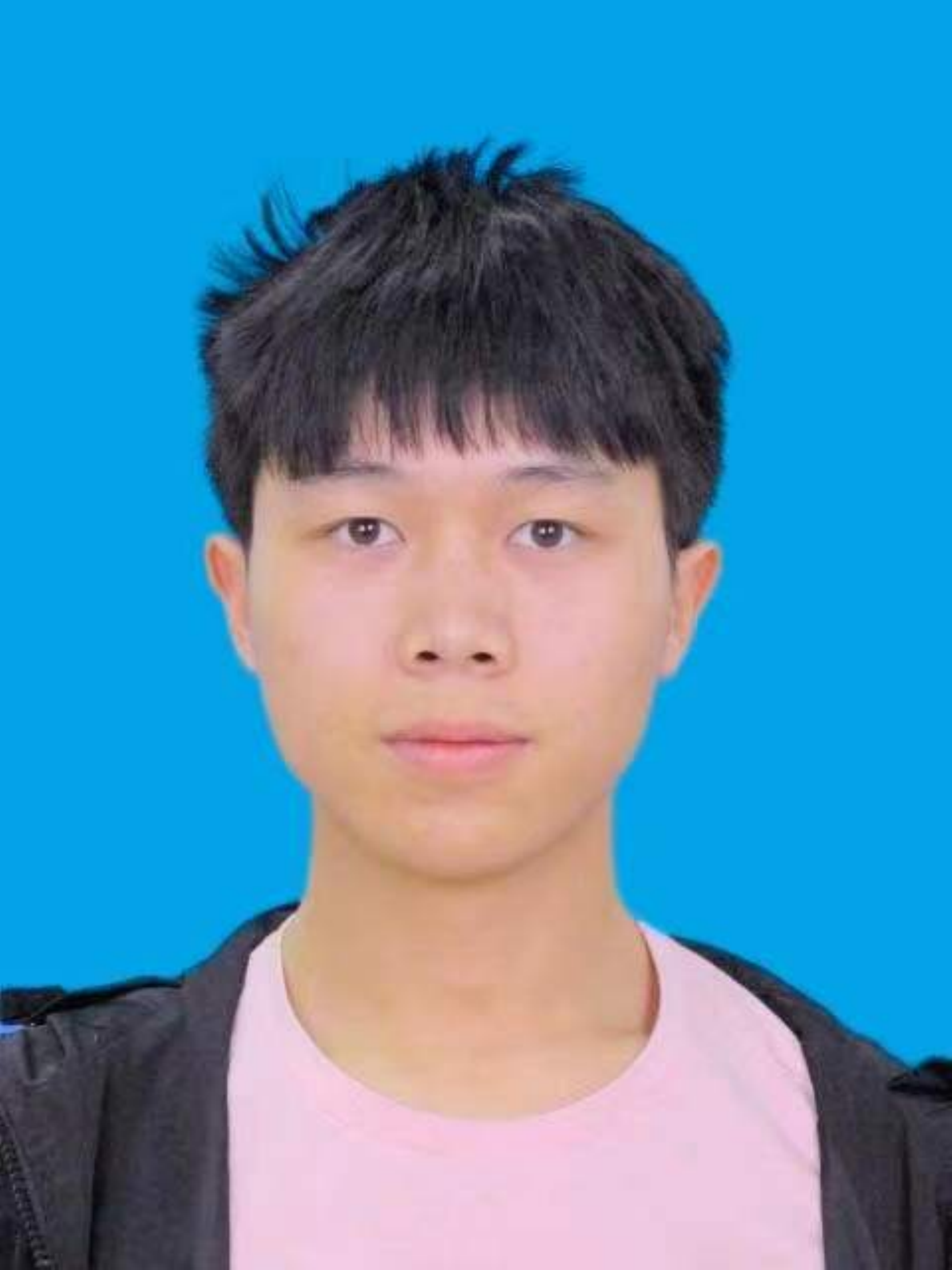}}]{Wei Tang}
received the bachelor's degree in engineering management from the Wuhan University of Science and Technology, Wuhan, China, in 2020. He is currently pursuing the master's degree with the China University of Geosciences. Wuhan, China. His research interest includes deep learning, point cloud processing and semantic segmentation.
\end{IEEEbiography}

\vspace{-33pt}
\begin{IEEEbiography}[{\includegraphics[width=1in,height=1.25in,clip,keepaspectratio]{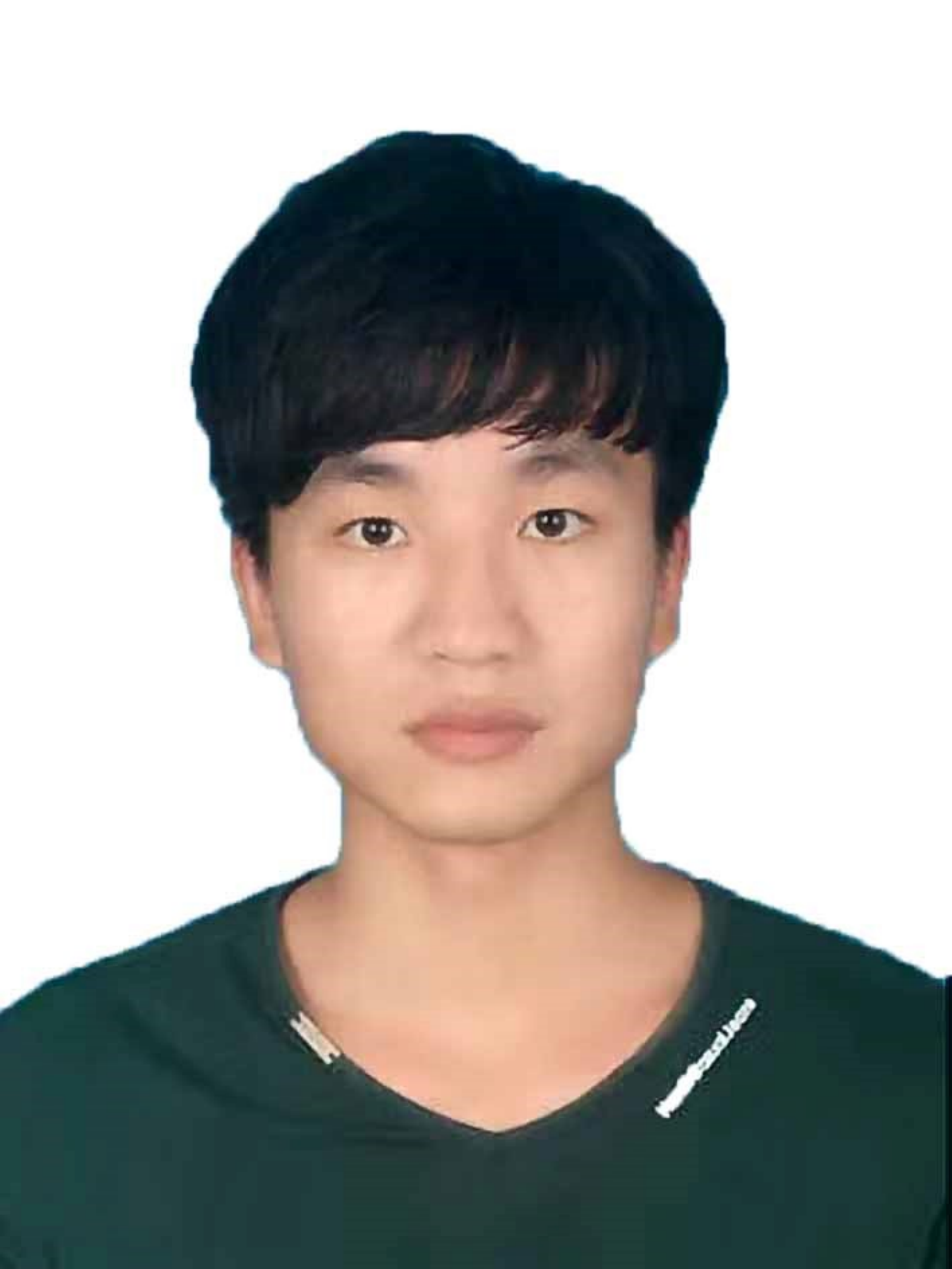}}]{Jie Wan}
received the M.S. degree in surveying and mapping engineering from China University of Geosciences, Wuhan, China, in 2020. He is currently a Ph.D candidate with the key laboratory of geological and evaluation of ministry of education, China University of Geosciences, Wuhan, China. His research interest includes deep learning, semantic segmentation and remote sensing image information extraction.
\end{IEEEbiography}

\vspace{-33pt}
\begin{IEEEbiography}[{\includegraphics[width=1in,height=1.25in,clip,keepaspectratio]{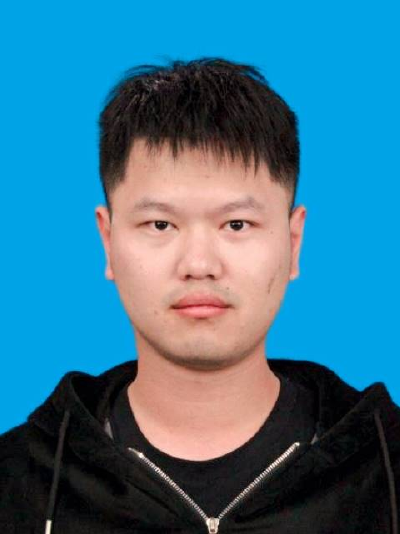}}]{Weichao Wu}
received a bachelor's degree in computer science and technology from China University of Geosciences Wuhan in 2019. He is currently pursuing the master's degree in software engineering from China University of Geosciences Wuhan. Since 2019, he has devoted himself to the research of computer vision, remote sensing and point cloud processing, especially the segmentation, completion and fusion of real scene point clouds and laser point clouds.
\end{IEEEbiography}

\end{document}